\documentclass{article}
\usepackage{spconf,amsmath,graphicx}
\usepackage{subfigure}
\usepackage{cite}
\usepackage{amsmath,amssymb,amsfonts}
\usepackage{tabularx,ragged2e,booktabs}
\usepackage{multirow}
\usepackage{tikz}
\usepackage{textcomp}
\usepackage{hyperref}
\usepackage{lipsum}

\newcommand{\norm}[1]{\left\lVert #1 \right\rVert}

\title{TOWARDS UNIVERSAL PHYSICAL ATTACKS ON CASCADED CAMERA-LIDAR 3D OBJECT DETECTION MODELS}
\name{{Mazen Abdelfattah} \quad   {Kaiwen Yuan} \quad  {Z. Jane Wang} \quad{Rabab Ward}}
\address{ECE Department, University of British Columbia\\
 	Vancouver, BC, Canada, V6T1Z4}
\begin{document}
%

\maketitle


%
\begin{abstract}
We propose a universal and physically realizable adversarial attack on a cascaded multi-modal deep learning network (DNN), in the context of self-driving cars. DNNs have achieved high performance in 3D object detection, but they are known to be vulnerable to adversarial attacks. These attacks have been heavily investigated in the RGB image domain and more recently in the point cloud domain, but rarely in both domains simultaneously - a gap to be filled in this paper. We use a single 3D mesh and differentiable rendering to explore how perturbing the mesh's geometry and texture can reduce the robustness of DNNs to adversarial attacks. We attack a prominent cascaded multi-modal DNN, the Frustum-Pointnet model. Using the popular KITTI benchmark, we showed that the proposed universal multi-modal attack was successful in reducing the model’s ability to detect a car by nearly 73\%. This work can aid in the understanding of what the cascaded RGB-point cloud DNN learns and its vulnerability to adversarial attacks.
\end{abstract}
\section{Introduction}
\label{sec:intro}

Most modern autonomous vehicles (AVs) employ cameras and LiDAR sensors to generate a complimentary representation of the scene in the form of dense 2D RGB images and sparse 3D point clouds. Using these data, deep neural networks (DNNs) have achieved state-of-the-art performance in 3D object detection. Such DNNs are however vulnerable to adversarial attacks (mainly in the image domain) that slightly alter the input but greatly affect the output. As safety is critical for self-driving cars, this paper investigates the vulnerabilities of cascaded multi-modal DNNs in 3D car detection.

\begin{figure}[htbp]
\centering
  \subfigure{
      \includegraphics[width=0.35\linewidth]{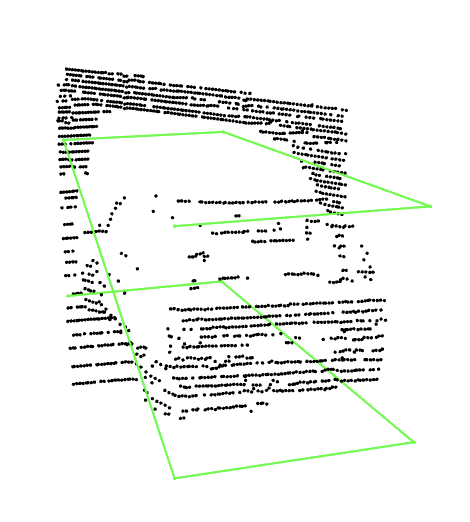}}
  \subfigure{
        \includegraphics[width=0.35\linewidth]{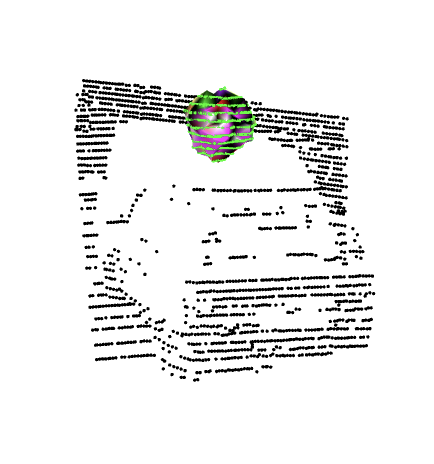}}
        \vspace{-0.3cm}
\caption{Illustrative example: Left: A car is detected with accurate bounding boxes. Right: The same car is not detected after adding an adversarial object with learned geometry and texture. The adversarial texture fools the 2D RGB detection pipeline, and the green points are the rendered LiDAR points added to the point cloud to fool the 3D point cloud pipeline.}
\label{fig1}
\end{figure}

Previous works demonstrated vulnerability of DNNs when the input is either an image \cite{adversarial , sharif2016accessorize} or a point cloud \cite{yang2019adversarial, Xiang_2019_CVPR}. Most point cloud attacks, simply alter or add individual points to the point cloud, and are not physically realizable due to properties of LiDAR sensors. More recently, a physically realizable adversarial attack on point clouds was made using a perturbed mesh that is placed in the 3D scene and rendered differentiably by simulating a LiDAR \cite{mich2019,  uber}. Existing attacks either target a single modality or are not physically realizable, limiting their applicability to real world scenarios.

Camera-LiDAR multi-model 3D object detectors can be divided into cascaded or fusion models. Cascaded models usually use an off-the-shelf 2D image detector to generate proposals that limit the 3D search space \cite{frustum, convnet, roarnet}. The points within the proposed regions are then fed into a 3D object detector for classification and bounding box regression. Fusion models extract and combine image and point cloud features in parallel using DNNs, and then a combined representation is sent to a detector for bounding box regression \cite{epnet,mvx}.

\begin{figure*}[t]
 \center
  \includegraphics[scale=0.4]{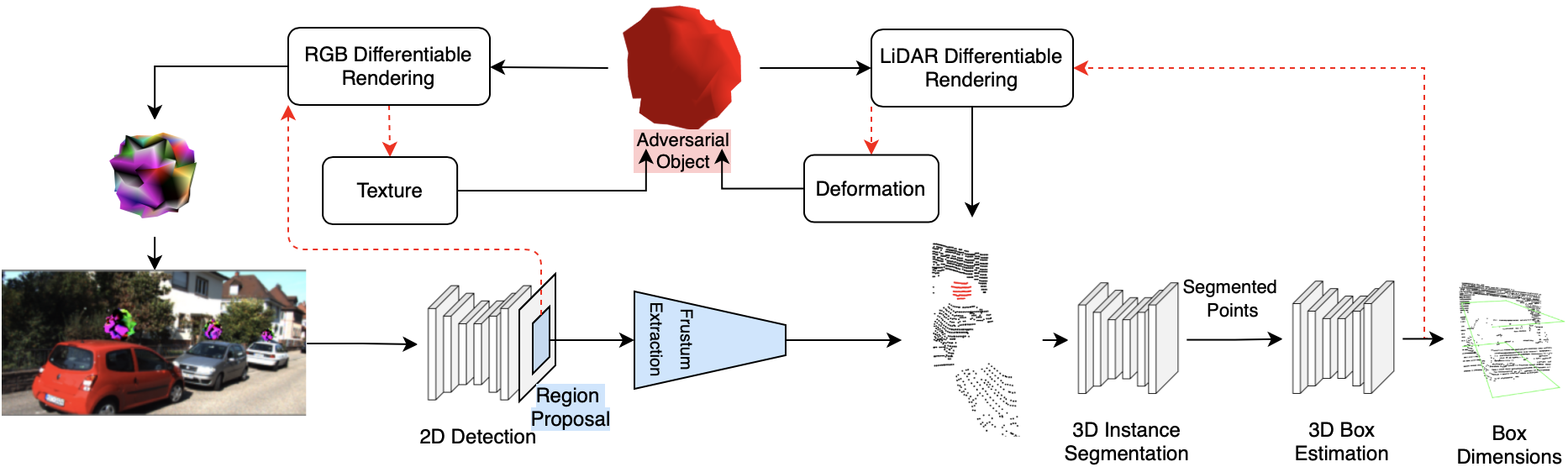}
  \vspace{-0.3cm}.
  \caption{The proposed attack pipeline on the cascaded F-PN model \cite{frustum}. Backpropagation is represented by a dashed red line.}
  \label{pipe}
\end{figure*}

This paper focuses on cascaded models because they are less computationally expensive and more interpretable than fusion models. They are therefore easier to train and deploy, especially that they also utilize mature 2D detection DNNs. While fused models are currently less studied, cascaded models are closer to industry deployment. Moreover, cascaded models have an implicit bias that in the post-proposal limited 3D point cloud an object must reside and so previous point cloud attacks could face a difficulty in attacking these models. Nonetheless, our proposed attack can extend easily to fusion models and other RGB-point cloud multi-modal DNNs. 

We propose a novel, universal, and physically realizable adversarial attack against a cascaded 3D car detection DNN. It attacks both image and point cloud. An adversarial 3D object is placed on top of a car in a 3D scene and then rendered to both point cloud and the corresponding RGB image by differentiable renderers. The shape and texture of the object are trainable parameters that are manipulated adversarially. To the best of our knowledge this is the first work that explores physically realizable multi-modal adversarial attacks on cascaded 3D object detection DNNs. This topic is critical because most self-driving vehicles now have LiDAR and camera sensors, and many advanced 3D detection models are multi-modal. Since the study of adversarial attacks on multi-modal DNN has so far been rare, this paper is towards addressing  this research gap. Based on evaluations on the KITTI dataset, our multi-modal attack reduced the average precision of car detection from 86\% to 28\% under easy conditions.

\section{Related Work}

Adversarial attacks are heavily explored in the image domain \cite{patch, adversarial, sharif2016accessorize}, starting with digital pixel-wise adversarial perturbations that were not realistic in the physical world \cite{adversarial}. Then constraints were established to ensure the physical feasibility of an attack and so patches \cite{patch}, and sunglasses \cite{sharif2016accessorize} were proposed to attack a model. Also, universal attacks were studied where an attack is not just trained on a single image  \cite{moosavi2017universal}. Recently, differentiable renderers were used to make adversarial attacks by altering the 3D geometry of an object or its lighting conditions and then rendering them to images for an adversarial attack on 2D detection DNNs \cite{zeng2019adversarial, liu2018beyond}. 

Our attack is very different from the aforementioned work: We aim not to learn a patch or a pixel value, but an adversarial texture on a 3D object that can attack a model even when viewed from many angles. Moreover, we target point cloud. The dense representation of RGB images made image attacks easier compared to point cloud attacks where the representation is very sparse. In the context of AVs, assuming a physically realistic attack on RGB images, it still does nothing with regards to a LiDAR sensor that is mainly concerned with 3D geometry. 

Early work on adversarial attacks on point cloud DNNs focused on perturbing point clouds by slightly moving, adding, or removing a few points \cite{ yang2019adversarial, Xiang_2019_CVPR}. These perturbations are largely theoretical, i.e. they have no guarantees that they can be realized physically. Also, these attacks were largely made on relatively dense point clouds that represent shapes and small objects, unlike the very sparse nature of AVs' point clouds.

To tackle the physical feasibility of a point cloud attack, \cite{mich2019} used an adversarial 3D mesh that is placed somewhere in a 3D scene around an AV. But it was only trained on a single example and an in-house dataset and so it was not universal, and their attack is not evaluated on a publicly available benchmark. To remedy these shortcomings, \cite{uber} proposed a universal attack on LiDAR using a mesh that is placed on top of a car, with the objective of minimizing the detection score. The attack was evaluated on different point cloud DNNs using the KITTI benchmark. All these approaches focus only on point cloud detection, while an AV also has a camera sensor.

\section{Methods}
We plan to learn a single adversarial object that is placed on top of a car in a 3D scene, with the goal that this adversarial object can significantly reduce the accuracy of a cascaded multi-modal DNN for 3D object detection. This object is rendered to point cloud and RGB image as if it was present in the original scene (see Fig.\hspace{4pt}\ref{pipe}). The consequences of such an attack are especially dangerous, because if those 2 main modalities are attacked, cars have very little recourse. Here we describe how this object is learned, the rendering methods, and the victim multi-modal DNN. 

\subsection{Attack method}
To make a realistic adversarial attack on multi-modal DNNs that can target both camera and LiDAR modalities, we require a representation that can maintain realistic 3D physical geometry and also can be differentiably rendered to RGB images and LiDAR. We choose a mesh for our adversarial object, since it provides an efficient representation of a 3D object and can be rendered to images or point clouds. This adversarial mesh is placed on top of a vehicle to avoid occlusion, and rotated around the vertical axis to have the same orientation as the vehicle.

To train the shape of the object, following previous work \cite{mich2019, uber}, we start with an initial mesh $S$ with $V$ vertices, where each initial vertex is defined as $v_i^0 \in \mathbb{R}^3$ $, i \in \{1, 2, ..., V\}$. We deform the shape of the mesh by displacing each vertex with a displacement vector $\hat{v_i}\in \mathbb{R}^3$, which is a learnable parameter to produce an adversarial mesh $S^{adv}$ with vertices $v_i$'s, as in Eqn. \eqref{deform}. We use a $4\times4$ transformation matrix $T$ to put the mesh on top of a car and give it the same orientation: 

\begin{equation}
    v_i = T \cdot (v_i^0 + \hat{v_i} ) \label{deform}
\end{equation}

For the texture of the adversarial mesh, we give each vertex a corresponding RGB color $c_i \in \mathbb{R}^3$, which is a learnable parameter. We then render the mesh to 2D and add it to the original RGB scene using a given projection matrix. To produce the texture of the mesh, each face is colored by the interpolation of the colors from the vertices surrounding it. 

To ensure our object is realistic, we put constraints on the size and smoothness of the mesh geometry. Also since we interpolate between vertex colors to produce the adversarial texture, the result is a realistic smooth texture without abrupt changes in color (a typical observation in RGB adversarial attacks using pixel-wise perturbations).

\subsection{Differentiable Rendering}
To realistically render a mesh to point cloud, we need to find which LiDAR rays would intersect with the mesh if it was placed in the original scene. We simulate the LiDAR used in capturing the dataset's point cloud by producing rays at the same angular frequencies and incorporating a small amount of noise that is present in this specific LiDAR. We then calculate the intersection points between these rays and our mesh's triangles in the 3D scene using the Möller–Trumbore intersection algorithm \cite{raycast}. Finally, for each ray with intersection points we take the nearest point, and add all the resulting points to the LiDAR point cloud scene. 

After placing our mesh on top of all cars in a 3D scene, we render the meshes to 2D using a given projection matrix. To train our adversarial texture, the rendering process needs to be differentiable. We therefore use the fast, differentiable rendering tools developed in \cite{ravi2020pytorch3d} to render our adversarial mesh from the 3D scene to RGB images, as shown in Fig. \ref{pipe}. 

\subsection{Victim Model} \label{AA}
As mentioned in section \ref{sec:intro}, we are attacking cascaded models. We chose the Frustum-PointNet (F-PN) model \cite{frustum} for many reasons. First, it's a pioneering work in the area of multi-modal DNN for AVs and many works were developed based on its mythology and architecture \cite{convnet, roarnet}. Moreover, it has a PointNet backbone which is commonly used in many single and multi-modal 3D detection DNNs, and so this attack could be a threat to other 3D detection DNNs that rely on PointNet \cite{pointnet}. Finally, it showed competitive results on the KITTI benchmark. 

As shown in Fig.\hspace{4pt}\ref{pipe}, F-PN first takes an RGB image through a 2D detection DNN, which proposes 2D bounding boxes. These bounding boxes are then projected to 3D space, thus producing frustum-shaped 3D search spaces that surround each object. Points within each frustum are extracted and sent through two PointNet based DNNs for 3D instance segmentation, and 3D box estimation. F-PN directly outputs one 3D bounding box estimation from a given point cloud frustum, since the assumption is that there is only a single object in a frustum. 

For image detection, we use YOLOv3 \cite{yolov3}. It is a standard fast 2D object detection DNN that outputs region proposals, classifications, and confidence scores. The F-PN and YOLOv3 models were pre-trained on the KITTI dataset. 

\subsection{Objective Functions}
We tried to attack the point cloud network in several ways and found that the most successful attack is achieved by minimizing the accuracy of the segmentation network. Assuming $N$ points in $M$ point clouds, the F-PN segmentation network outputs 2 logits for each point, estimating whether it belongs to an object or not,  $logit_i \in \mathbb{R}^2, i \in \{1, 2, ..., N\}$. We softmax these logits and extract the highest probability given to a ``car" point, $p_k = \max\limits_{j} \{ softmax(logit_j)\}$ in a point cloud $k$, where $j$ is a point classified as car. To attack the 3D segmentation network, we train the shape of the mesh to minimize this probability. Finally, following previous work \cite{uber}, we weighed the objective function by the IoU score between the ground truth $y_{gt}$ and predicted $\hat{y}$ bounding boxes:

\begin{equation}
\mathcal{L}_{mesh} = \sum_{k\in M} -1 * log(1-p_k)*IoU(y_{gt},\hat{y}) \label{loss_mesh}. 
\end{equation}
As for the image loss function, we simply minimize the ``objectness" score given to resulting car classifications.

We also want to ensure that the geometry of the mesh is smooth and realistic, so we minimize a Laplacian loss \cite{liu2019soft}: $\mathcal{L}_{lap}=\sum_i\norm{\delta_i}^2_2 \label{lap}$, where $\delta_i$ is the distance between a vertex $v_i$ and the centroid of its neighbors $\delta_i = v_i -  \frac{1}{\norm{N(i)}}\sum_{j\in N(i)}v_j$. 

The overall point cloud attack objective function is:

\begin{equation}
\mathcal{L}^{pc}_{adv} = \mathcal{L}_{mesh} + \lambda \mathcal{L}_{lap} \label{loss}
\end{equation}

\begin{figure}[t]
\centerline{\includegraphics[scale=0.44]{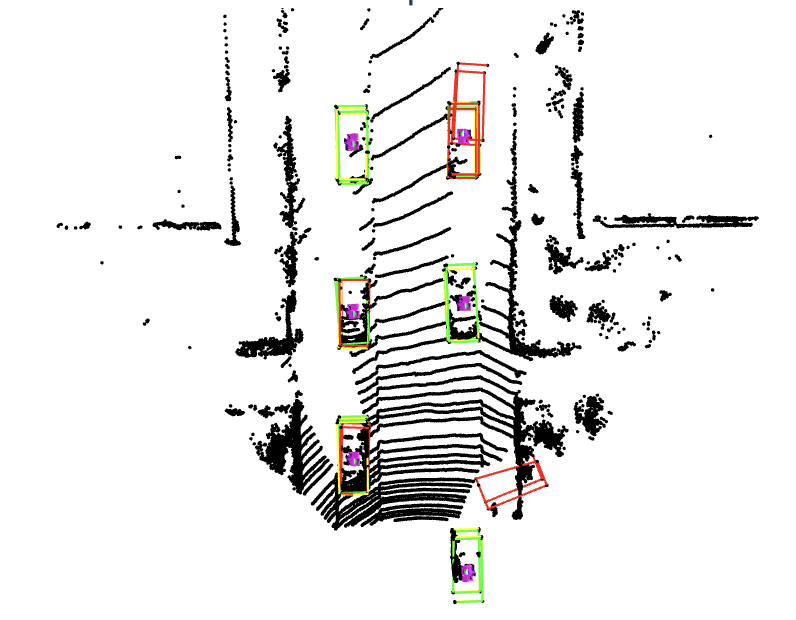}}
\caption{An example of the RGB+point cloud attack. Green boxes are ground truth, yellow boxes are detections without adversarial attack, and red ones are detections after the adversarial attack.}
\label{res}
\end{figure}

\vspace{-0.3cm}
\section{Experiments}
\subsection{Experiment Setup}
We use the KITTI dataset \cite{geiger2012we}, a popular benchmark for 3D detection in autonomous driving. Roughly half of the samples are used for training and the other half for validation. All reported results in the following section are from the validation set. We use ground truth 2D bounding boxes to generate the frustums used in training and evaluating the attack on the point cloud pipeline. We use YOLOv3 generated 2D bounding boxes to measure the effect of an image-only attack and image + point cloud attack on detection. This is a universal attack, i.e., the shape and texture of the object are trained on the entire training set.

We first perturb the geometry of the object, to attack the 3D part. We start with a 40 cm radius isotropic sphere mesh with 162 vertices and 320 faces. We apply box constraints to keep the mesh's 3 dimensions below the initial levels. An ADAM optimizer is then used to iteratively deform the mesh to minimize \eqref{loss}. Once we have an adversarial shape we render it to 2D images and train a universal adversarial texture. We give each vertex an initial color and then we use an ADAM optimizer to learn which color, for each vertex, would produce an adversarial texture to attack the RGB pipeline.

\vspace{-0.1cm}
\subsection{Results \& Discussion}
We measure the success of an attack by the reduction in the average precision (AP) of car detection due to the introduction of an adversarial mesh to the scene. Table \ref{pc} shows the bird's eye view (BEV) AP results of our trained F-PN after different types of attacks. We show the results for 3 difficulty levels (based on occlusion) with an IoU threshold of $0.7$. 

\begin{table}[htbp]
\begin{center}
\begin{tabularx}{\columnwidth}{ l|XXX }
\toprule
\textbf{Attack Type$^{\mathrm{a}}$}&\textbf{Easy} & \textbf{Moderate} &\textbf{Hard} \\
\midrule
\textbf{\textit{No Attack}} & 85.66 & 83.62 & 76.23\\

\textbf{\textit{PC: Adv Shape}} & 37.36 & 37.79 & 39.10  \\

\textbf{\textit{Img: Adv Texture}} & 51.85 & 35.98 & 31.49 \\

\textbf{\textit{PC + Img: Adv Object}} & \textbf{27.50} & \textbf{22.67} & \textbf{19.21} \\
\bottomrule
\multicolumn{4}{l}{$^{\mathrm{a}}$Attack on the point cloud (PC) or image (Img) pipeline.} 
\end{tabularx}
\end{center}
\vspace{-0.1cm}
\caption{BEV 3D car detection AP results}
\label{pc}
\end{table}

It can be challenging to attack the point cloud part of a cascaded model for several reasons. First, these models are built with an implicit bias that an object must reside in a proposed region and so they can be resistant to simple perturbations. Also, the post-proposal point cloud is very sparse and small in size. Thus there is not much space for an adversarial attack to take place. Nonetheless, our adversarial mesh was able to reduce the localization BEV AP by nearly 45\% by just targeting the point cloud pipeline. This can be attributed to cascaded models' limited exposure to examples where objects could be on cars during training. The frustum limits the 3D search space by focusing only on the body of a car. This is a weakness in cascaded models, since the limited 3D search space means that it may not learn at all about the surrounding 3D scene. This poses a grave danger for autonomous driving since the dimensions and locations of surrounding vehicles and objects contribute heavily to decision making. As shown in Fig \ref{res}, our adversarial object not only led to some cars going entirely undetected, but also introduced other detection errors.

The RGB pipeline is the bottleneck of detection performance in camera-LiDAR cascaded models. The 2D proposals decide exactly where to search for objects and so if that fails the entire model fails. Using 2D region proposals doesn't utilize a main purpose behind LiDAR which is to avoid cases where lighting and occlusion affect localization and detection. RGB attacks are much simpler and less computationally expensive than point cloud attacks, and they're more easily reproduced in real life which can make them more dangerous. 

Many previous work showed that data augmentation can make a DNN more robust to adversarial attacks. In the training of our network, we included the data augmentation step in the original work \cite{frustum} by randomly scaling and shifting the proposed 2D bounding boxes. This gave the network certain robustness, as shown in the relatively high AP of car detection even after the RGB attack under the Easy case. Nonetheless, when occlusion increased, detection accuracy rapidly decreased. Under moderate circumstances, BEV AP decreased from 83.62\% to 22.67\%, a nearly 73\% reduction. 

\vspace{-0.2cm}
\section{Conclusion}
We proposed a novel, universal, and physically realistic adversarial attack on a cascaded camera-LiDAR DNN for 3D object detection. We manipulated mesh geometry and texture and used differentiable rendering to study the vulnerability of cascaded DNNs. We found that cascaded models are vulnerable to adversarial attacks and that detection accuracy can decrease by nearly 73\% if we attack both camera and LiDAR modalities simultaneously. While we have applied here the proposed attack to a multi-modal cascaded model, it can be extended in future work to investigate the vulnerability of different multi-modal DNN models (e.g., fusion models) for different tasks.


\bibliographystyle{IEEEbib}
\bibliography{ref}

\end{document}